\documentclass[conf]{new-aiaa}
\usepackage[utf8]{inputenc}

\usepackage{amsthm}

\usepackage{adjustbox}

\usepackage{amsmath}
\usepackage[version=4]{mhchem}
\usepackage{siunitx}
\usepackage{longtable,tabularx}
\usepackage[table]{xcolor}

\usepackage{mathrsfs}
\usepackage{float}

\usepackage{pgfplots}
\usepackage{color, colortbl}
\definecolor{Gray}{rgb}{0.85,0.85,0.85}
\definecolor{LtGray}{rgb}{0.95,0.95,0.95}

\newcommand{\norm}[1]{\left\lVert#1\right\rVert}

\usepackage{amsmath} 
\DeclareMathOperator*{\argmax}{arg\,max}

\usepackage{hyperref}       
\usepackage{url}            
\usepackage{booktabs}       
\usepackage{amsfonts}       
\usepackage{nicefrac}       
\usepackage{microtype}      
\usepackage{color}
\usepackage{csquotes}

\usepackage{graphicx}
\graphicspath{ {images/} }
\usepackage{scrextend}
\usepackage{booktabs}
\usepackage{float}
\usepackage{listings}

\usepackage{tikz}
\usetikzlibrary{math}
\usetikzlibrary{shapes,arrows}
\usetikzlibrary{positioning}
\usetikzlibrary{arrows.meta}
\usetikzlibrary{decorations}
\usetikzlibrary{decorations.pathmorphing}
\usetikzlibrary{shadows}
\usetikzlibrary{arrows.meta,
                calc, chains,
                quotes,
                positioning,
                shapes.geometric}
                
\usepackage{times}
\usepackage{tkz-euclide}
\usetkzobj{all}

\usepackage{subcaption}

\usepackage{algorithm}
\usepackage{algorithmicx}
\usepackage[noend]{algpseudocode}

\usepackage{relsize}

\usepackage{titlesec}
\titlespacing*{\section}{0pt}{1em}{1em}
\titlespacing*{\subsection}{0pt}{1em}{1em}

\makeatletter
\def\BState{\State\hskip-\ALG@thistlm}
\makeatother

\tikzstyle{block} = [draw,fill=blue!20,minimum size=2em]

\tikzstyle{branch}=[fill,shape=circle,minimum size=3pt,inner sep=0pt]

\title{Scalable FastMDP for Pre-departure Airspace Reservation and Strategic De-conflict}

\author{
  Joshua R. Bertram    \quad    Joseph Zambreno \\
  Iowa State University\\
  Ames, IA 50011 \\
  \texttt{\{bertram1, zambreno\}@iastate.edu} \\
  \normalfont{Peng Wei} \quad \\
  George Washington University\\
  Washington, DC 20052 \\
  \texttt{pwei@gwu.edu} \\
}

\definecolor{gridcolor}  {RGB}{  2, 82,163}  
\definecolor{blockcolor1}{RGB}{161,195,236}  
\definecolor{blockcolor2}{RGB}{181,215,236}
\definecolor{threadcolor}{RGB}{235,235,235} 

\definecolor{textcolor}  {RGB}{255,255,255}  

\begin{document}

\maketitle

\begin{abstract}
    Pre-departure flight plan scheduling for Urban Air Mobility (UAM) and cargo delivery drones will require on-demand scheduling of large numbers of aircraft.  We examine the scalability of an algorithm known as FastMDP which was shown to perform well in deconflicting many dozens of aircraft in a dense airspace environment with terrain.  We show that the algorithm can adapted to perform first-come-first-served pre-departure flight plan scheduling where conflict free flight plans are generated on demand.  We demonstrate a parallelized implementation of the algorithm on a Graphics Processor Unit (GPU) which we term FastMDP-GPU and show the level of performance and scaling that can be achieved.  Our results show that on commodity GPU hardware we can perform flight plan scheduling against 2000-3000 known flight plans and with server-class hardware the performance can be higher.  We believe the results show promise for implementing a large scale UAM scheduler capable of performing on-demand flight scheduling that would be suitable for both a centralized or distributed flight planning system.
\end{abstract}

\section{Introduction}

\begin{figure}[b]
\centering
\begin{subfigure}[t]{3in}
\includegraphics[width=3in]{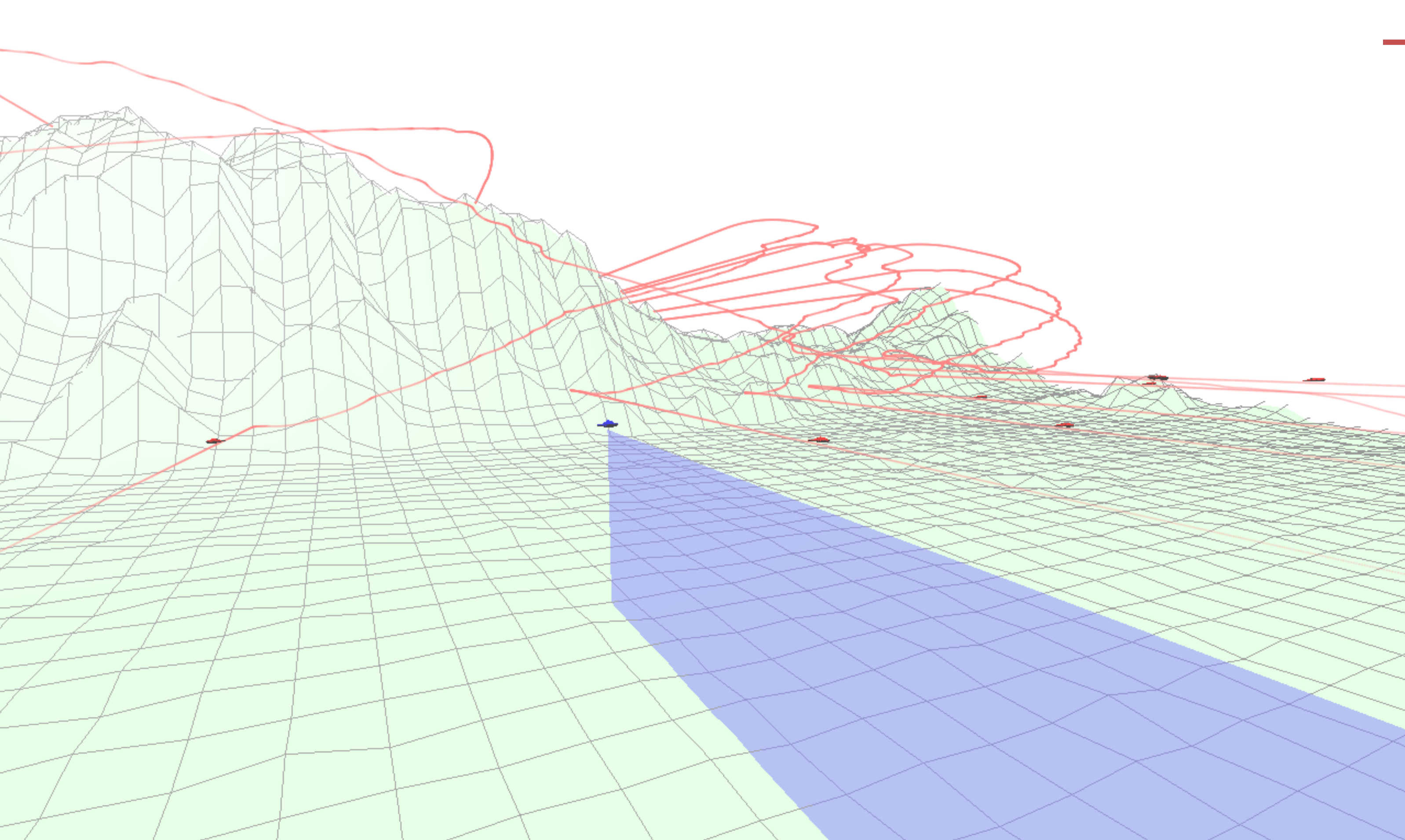}
\end{subfigure}
\quad
\begin{subfigure}[t]{3in}
\includegraphics[width=3in]{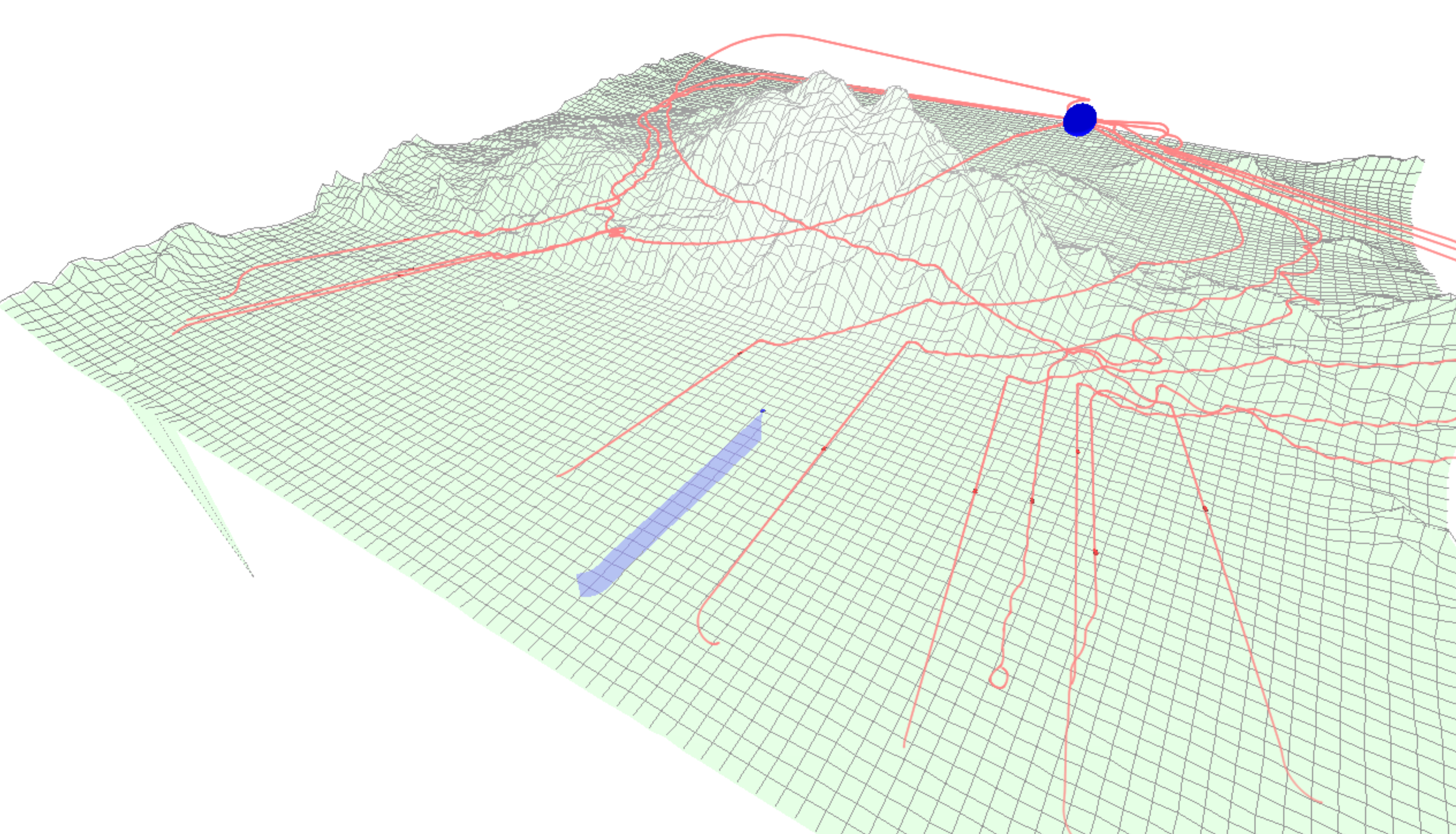}
\end{subfigure}
\caption{Agent (blue aircraft and blue flight track) flying through an airspace with known flight plans of other aircraft (red aircraft and red trajectories).}
\label{flightplans}
\end{figure}

Urban Air Mobility (UAM) is an envisaged air transportation concept, in which intelligent flying machines will safely and efficiently transport passengers and cargo within urban areas by rising above traffic congestion on the ground. Aircraft companies such as Boeing, Airbus, Bell, Embraer, Joby, Kitty Hawk, Pipistrel, and Volocopter are building and testing electric vertical take-off and landing (eVTOL) aircraft to ensure UAM becomes an integral part of the daily life \cite{UBERElevate2016fast}. Meanwhile, in order to make UAM operations scalable, new airspace management concepts for highly dynamic and dense air traffic are being studied by NASA, FAA, Uber, Airbus, etc \cite{NASAUrbanAirMobility}\cite{UberDynamicSkylanes}\cite{airbus2017uam}. Furthermore, NASA’s UAM Grand Challenge was announced to engage the UAM community and promote public confidence through a series of system level safety and integration scenarios \cite{NASAGrandChallenge}.

A major challenge to scale the UAM operations and accommodate high-density urban air traffic is strategic de-conflict for eVTOL flights. ``Strategic" here refers to pre-departure, whose procedure has to be completed while the aircraft is still on the ground, instead of tactical real-time conflict resolution or collision avoidance. The essence of ``strategic de-conflict" is how to reserve airspace volume in a safe and efficient manner \cite{zhu2016low}\cite{zhu2019pre}. In this work, we focus on efficient and scalable algorithms for pre-departure airspace reservation in free flight strategic de-conflict. We assume all the airspace volume in a certain region (except the obstacles or restricted area) can be occupied by airspace users, instead of a structured airspace with routes and waypoints.

Our contribution of this paper is that this is the first scalable algorithm for pre-departure airspace reservation and strategic de-conflict to support free flight UAM operations. We respect the airspace user fairness by following the first-come-first served (FCFS) manner. The algorithm offers great flexibility for implementation. It can be run by either a fleet dispatch center (e.g. Uber), a personal user (personal owned eVTOL pilot), or by the airspace manager (e.g. the FAA). The algorithm can handle both by-scheduled UAM operations (processing flight plans or airspace reservation requests in batches) or air taxi like operations (processing flight plans or airspace reservation requests one by one). Our framework exploits both route and take-off time to achieve strategic de-conflict quickly and efficiently. The algorithm provides a recommended flight plan, which contains not only the recommended route (a series of dynamic airspace volume reservations), but also the recommended take-off time from several candidates.  \textit{Note to reviewers:  We plan on adding the departure time delay to the implementation over the summer and fall before the final paper deadline.}


The approach used in this paper formulates the problem as a Markov Decision Process (MDP) and then uses an algorithm known as FastMDP to quickly and efficiently solve the MDP to generate a conflict free trajectory.  A request is submitted to the system containing an aircraft identifier, a source location, and a destination location and a conflict free flight path is returned to the user.  When a conflict free flight path is generated, it is stored by the system as an ``accepted'' flight plan in a database of accepted flight plans.  Any future requests will consider both terrain and all previously accepted flight plans when generating new conflict free flight plans.  Requests can consist of a single aircraft, or can consist of batches of aircraft.  
The implementation of the algorithm described in this paper takes advantage of parallelization inherent in the problem to create a massively parallel implementation which scales to thousands of aircraft on commodity Graphics Processing Units (GPUs).  This implementation is used to study the level of scalability that can be achieved by the algorithm on different classes of hardware.  While the algorithm is capable of running on both server class GPU hardware and embedded class GPU boards suitable for low size-weight and power applications, the focus of this paper is on GPU hardware that is too large and power hungry for putting on board a typical aircraft.

The paper is structured as follows: Section II contains related work, Section III contains background material on Markov Decision Processes (MDPs) and the FastMDP algorithm, Section IV describes the method used in this paper, Section V describes the experimental setup, Section VI describes the experimental results, and the paper closes with Section VII where the conclusion and future work are discussed.

\section{Related Work}

There have been many important contributions to the topic of guidance algorithms with collision avoidance capability for small unmanned aerial aircraft. 

Many papers apply different techniques to manage aircraft from a centralized controller by formulating the problem as an optimal control problem.
These methods can be based on semidefinite programming \cite{frazzoli2001resolution}, nonlinear programming \cite{raghunathan2004dynamic,enright1992discrete}, mixed integer linear programming \cite{schouwenaars2001mixed,richards2002aircraft,pallottino2002conflict,vela2009mixed}, mixed integer quadratic programming \cite{mellinger2012mixed}, sequential convex programming \cite{augugliaro2012generation,morgan2014model}, second-order cone programming \cite{acikmese2007convex}, evolutionary techniques \cite{delahaye2010aircraft,cobano2011path}, and particle swarm optimization \cite{pontani2010particle}.
Besides formulating this problem using optimal control framework, methods such as visibility graph \cite{hoffmann2004stanford} Voronoi diagrams \cite{howlet2004practical}, and $A^*$ search \cite{meng2010uav,xia2009path} can also handle the path planning problem for aircraft. These methods work well for 2D or 3D waypoint planning, but fail to scale when aircraft dynamics are considered. To address this issue, sample-based planning algorithms are proposed, such as probabilistic roadmaps (PRM) \cite{kavraki1994probabilistic}, rapidly-exploring random trees (RRT) \cite{lavalle1998rapidly}, and RRT* \cite{karaman2011sampling}.
Model Predictive Control \cite{shim2007evasive, shim2003decentralized} can be used to solve collision avoidance problem but the computation load is relatively high. 
Potential fields \cite{sigurd2003uav, langelaan2005towards} are computationally fast, but in general they provide no guarantees of collision avoidance. 
Machine learning and reinforcement learning based algorithms \cite{kahn2017plato, zhang2016learning, ong2016markov, chen2017decentralized} have promising performance, but typically require a long offline training time.  Monte Carlo Tree Search (MCTS) algorithm \cite{yang2018autonomous} does not need time to train before the flight and can provide an ``anytime'' result depending on the available computation time, but the aircraft can only adopt several discretized actions at each time step. A geometric approach \cite{han2009proportional,park2008uav, krozel2000decentralized, van2011reciprocal} can be also applied for the collision avoidance problem and the computation time only grows linearly as the number of aircraft increases. 
DAIDALUS (Detect and Avoid Alerting Logic for Unmanned Systems) \cite{munoz2015daidalus} is another geometric approach developed by NASA. The core logic of DAIDALUS consists of: (1) definition of self-separation threshold (SST) and well-clear violation volume (WCV), (2) algorithms for determining if there exists potential conflict between aircraft pairs within a given lookahead time, and (3) maneuver guidance and alerting logic. The drawback of these geometric approaches is that it can not look ahead for more than one step (it only pays attention to the current action and does not take account of the effect of subsequent actions) and the outcome can be locally optimal in the view of the global trajectory.


There are many well known methods for solving MDPs including value iteration and policy iteration, which are iterative methods based on the dynamic programming approach proposed by Bellman \cite{bellman2013dynamic}.  These algorithms use a table-based approach to represent the state-action space exactly and iteratively converge to the optimal policy $\pi^*$ and corresponding value function $V^*$.  These table-based methods have a well known disadvantage that they quickly become intractable.  As the number of states and actions increases in number or dimension, the number of entries in the table increases exponentially.  Many real world problems quickly exhaust the resources of even high performing computers due to the well known `curse of dimensionality' \cite{bellman2013dynamic}.
Many attempts have been made to allow MDPs to scale to larger problems.  Factored MDPs \cite{schuurmans2002direct,guestrin2003efficient}  attempt to alleviate the problem of state space explosion by identifying subsets of the MDP that can be broken into smaller problems.  Approximation methods under the general umbrella of Approximate Dynamic Programming have been used as a compromise to obtain reasonable approximations of the underlying true value function in cases where the state-action space (or the transition matrix $T$) is too large to represent with traditional exact methods, which are summarized by the \cite{bertsekas1995dynamic,powell2007approximate}.  
Notably, linear function approximation methods such as GradientTD methods \cite{gradientTD,tdcandgtd2,precup2001off}, statistical evaluation methods such as Monte Carlo Tree Search \cite{kocsis2006bandit}, and non-linear function approximation methods such as TDC with non-linear function approximation \cite{bhatnagar2009convergent} and DQN \cite{DQNatari} are good examples of some of the approaches taken using approximation.  

Traffic Management Initiatives (TMIs), including Ground Delay Program, Airspace Flow Program and Collaborative Trajectory Options Program, are set of tools air traffic managers use to balance air traffic demand with airspace capacity \cite{odoni1987flow, Brennan2007AFP, ZhuPhDThesis}.  The essence of these programs is to apply ground delay or assign longer route to flights who would otherwise experience more expensive and unsafe air delay. 
Within the area of pre-departure flight planning, \cite{zhu2016low} defines a moving dynamic geofence around an aircraft during its flight.  The dynamic geofence represents a safety margin around the aircraft which is sufficiently large to guarantee safe separation.  The dynamic geofence is formulated as a convex polyhedra and intersection tests between polyhedra are used to detect conflicts with other flight plans.  Network optimization is used to find feasible paths between a grid of waypoints representing the possible paths the aircraft can take when navigating between points in an urban area.  Experiments show an airspace utilization improvement of 70-80\% over a reserved corridor around the flight plan for the duration of the flight.  
This concept is expanded upon in \cite{zhu2019pre} where the dynamic geofence is used, but the solution is formulated as a two-level linear programming problem.  The first level of the problem is a discretized version of the problem which resolves scheduling conflicts using integer programming.  The second level performs speed profile smoothing using linear programming on the discretized solution from the first level.  While the method provides a global optimum, no simulation or numerical results are provided for the runtime for sample problem sizes.  
In \cite{bosson2018simulation} NASA explores and extension to AutoResolver for UAVs to model realistic air traffic management scenarios in the Dallas-Fort Worth metroplex, studying loss of separation and resolutions in a simplified structured airspace, and shows AutoResolver can effectively introduce ground delays and alter fixed flight paths along the structured airspace routes to avoid conflicts.  Performance is shown in terms of the impacts to flight schedules, though performance time to run the algorithm itself was not reported.  
In \cite{guerreiro2019mission}, NASA explores the use of Mission Planner which performs pre-departure flight planning for UAVs / UAM in a first-come-first-served manner.  Mission Planner models both the network routing and trajectory generation problems.  For a given set of goods or people that need moved through the network, Mission Planner takes into account suitable aircraft availability, vertiport capacity, and generates a set of flights that will satisfy the demand.  It then builds candidate flight paths and iteratively uses a set of resolution strategies to resolve conflicts or constraint violations that are discovered through the duration of each flight, resulting in a viable, conflict free flight path.  Performance of the algorithm is examined in terms of scheduling a random set of realistic flights over a 3 hour window, where the number of flights was randomly sampled to be from 1000 to 10000 distributed over the 3 hour time window.  A study is performed on the effectiveness of each resolution strategy.  No results are provided for the run time of the algorithm itself.

\section{Background}

\begin{figure}[tbp]
\centering
\begin{subfigure}[t]{3in}
\begin{adjustbox}{scale={0.65}{0.75}, padding=0 0ex 0 0}
\begin{tikzpicture}[decoration={markings, mark=at position 1.0 with {\arrow{>}};}]
      
      
      \tkzDefPoint(5,-.05){A}
      \tkzDefPoint(5,-1){B}
      \tkzLabelPoint[above](A){$s_i$}
      \tkzDrawSegments[postaction={decorate}](A,B)
      
      
      
      
      \tkzDefPoint(6.5,1.20){A}
      \tkzLabelPoint[above](A){$r$}
      \tkzDefPoint(3.5,1.20){A}
      \tkzLabelPoint[above](A){$r$}
      \tkzDefPoint(5.3,1.00){A}
      \tkzDefPoint(8,1.0){B}
      \tkzDrawSegments[postaction={decorate}](A,B)
      \tkzDefPoint(4.7,1.00){A}
      \tkzDefPoint(2,1.0){B}
      \tkzDrawSegments[postaction={decorate}](A,B)

      \draw[->] (0 ,0) -- (10,0) node[right] {$state$};
      \draw[->] (0 ,-3) -- (0,3) node[above] {$value$};
      \draw[thick,domain=0:10,samples=1000,variable=\x,blue] plot ({\x},{ -2 * 0.7^(abs(\x-5)) * (\x<8) * (\x>2) });
      \draw[thick,dotted,domain=0:10,samples=1000,variable=\x,blue] plot ({\x},{ -2 * 0.7^(abs(\x-5))  });
     
      \node at (0,-4.5) {};  
\end{tikzpicture}
\end{adjustbox}
\caption{A risk well showing exponential decay of a negative reward out to a fixed radius beyond which the negative penalty is truncated.}
\label{riskwell}
\end{subfigure}
\quad
\begin{subfigure}[t]{3in}
\centering
\includegraphics[width=3in]{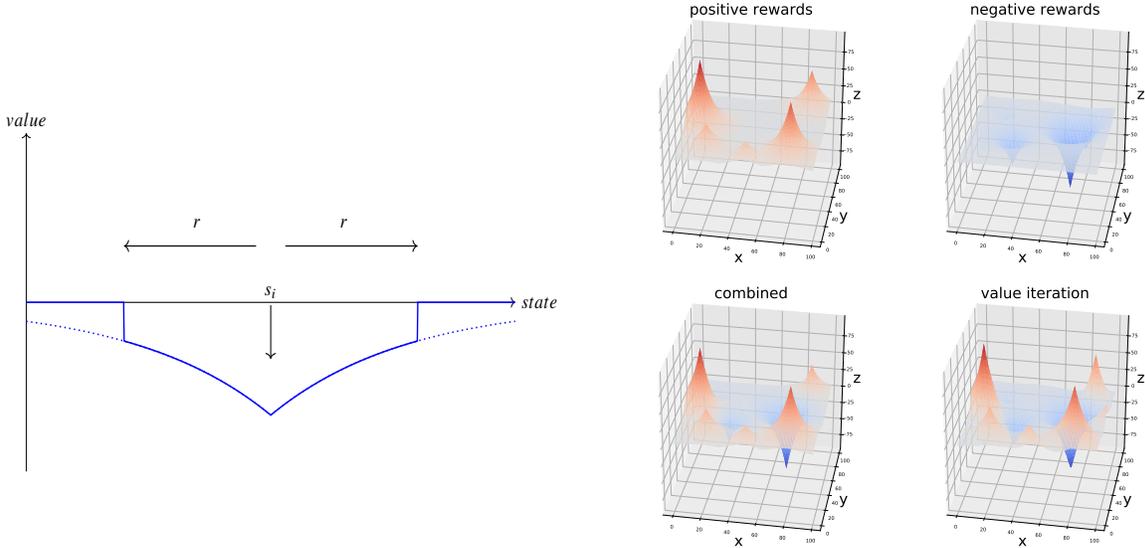}
\caption{FastMDP solving positive and negative rewards, combining the results, and comparing to the solution produced by the value iteration algorithm traditionally used to solve MDPs.}
\label{compare}
\end{subfigure}
\caption{FastMDP solves MDP using peaks that represent positive and negative rewards}
\end{figure}

    Markov Decision Processes (MDPs) are a framework for decision making with broad applications to finance, robotics, operations research and many other domains \cite{suttonbarto}.  MDPs are formulated as the tuple $(s_t, a_t, r_t, t)$ where $s_t \in S$ is the state at a given time $t$, $a_t \in A$ is the action taken by the agent at time $t$ as a result of the decision process, $r_t = R(s_t, a_t)$ is the reward received by the agent as a result of taking the action $a_t$ from $s_t$ and arriving at $s_{t+1}$, and $R(s_t, a_t)$ is known as the \emph{reward function}.  The dynamics of the environment are described by the transition function $T(s_t, a, s_{t+1})$ and capture the probability $p( s_{t+1} | s_t, a_t )$ of transitioning to a state $s_{t+1}$ given the action $a_t$ taken from state $s_t$.  
    A policy $\pi$ can be defined that maps each state $s \in S$ to an action $a \in A$.  From a given policy $\pi \in \Pi$ a value function $V^\pi(s)$ can be computed that computes the expected return that will be obtained within the environment by following the policy $\pi$.  We use the \emph{infinite horizon discounted reward formulation} where a parameter $\gamma \in (0, 1)$ is defined which is applied at each step to determine return.  A small value of $\gamma$ favors short term reward versus long term reward, whereas a large value of $\gamma$ near $1.0$ favors long term reward versus short term reward.

    The solution of an MDP is termed the optimal policy $\pi^*$, which defines the optimal action $a^* \in A$ that can be taken from each state $s \in S$ to maximize the expected return.  From this optimal policy $\pi^*$ the optimal value function $V^*(s)$ can be computed which describes the maximum expected value that can be obtained from each state $s \in S$.  And from the optimal value function $V^*(s)$, the optimal policy $\pi^*$ can also easily be recovered.

    We refer to the path taken through the state space as a result of following the optimal policy as the \emph{optimal trajectory}.  We define the UAV planning problem as finding the optimal trajectory through the space such that the UAV maximizes its future expected reward.  From any starting state, by following the optimal policy $\pi^*$, we are guarantees to also follow the optimal trajectory.
    MDPs are interesting because their solution simultaneously provides the optimal action $a^*$ to perform from every state and can be viewed as analogous to a vector field in a continuous space.
    
A challenge with traditional MDP solution methods is that they often take a great deal of time to solve due to the iterative nature that is required to solve them.  One also finds that as the number of states or actions grows, the amount of time or memory required to solve the MDP grows exponentially leading to issues of intractability.  This is somewhat mitigated by the use of approximation methods which lead to tractable solution methods for MDPs which, while iterative, lead to solutions within for many interesting problems on time scales ranging from seconds to days depending on the particular problem.  MDPs are typically not suitable for real-time applications, though there are some on-line methods which allow MDPs to be solved without an explicit pre-training phase.

The FastMDP algorithm proposed in \cite{bertramMastersThesis} represents a radical departure from the traditional approach to solving MDPs.  FastMDP solves a certain useful subclass of MDPs much more quickly than traditional methods by taking advantage of structure within the value function.
FastMDP relies on the observation that positive and negative rewards in an MDP can be described as exponentially decaying \emph{peaks} in the value function which can be combined in a particular way to reconstruct the value function. 

In \cite{bertramMastersThesis}, a method is described to combine the positive and negative peaks together such that they closely approximate the value function produced by solving a MDP using traditional methods, as shown in Figure \ref{compare}.  For UAV collision avoidance problems in \cite{bertramMastersThesis, bertram2020distributed, bertram2020efficient} positive rewards are modeled as exponentially decaying peaks while negative rewards are modelled as \emph{risk wells} which decay exponentially out to a fixed radius, where they are then truncated.  
Risk wells capture the idea that if a penalty is far enough away from an agent's current position it can be safely ignored, while also encoding that the closer the agent is to a negative reward the riskier it is to be near that reward.
As described in \cite{bertramMastersThesis}, the risk well formulation also has the advantage that it can be processed using the same efficient algorithm that is used to process positive rewards.  This leads to a very efficient way to model UAV collision avoidance problems that has been successfully demonstrated to solve interesting, practical problems.

\begin{algorithm}[tbph]
    \footnotesize
    \caption{CPU based FastMDP algorithm from \cite{bertram2020distributed}} 
    \label{orig_algo}
    \begin{algorithmic}[1]
    \Procedure{DistributedUam}{$\textit{aircraftState}, ~ \textit{worldState}$}
        \State $\mathbf{S_t}  \gets \mathbf{S}_0$ // randomized initial aircraft states
        \State $\mathbf{A} \gets$ \textit{aircraft actions (precomputed)}
        \State $\mathbf{L} \gets$ \textit{aircraft limits (precomputed)}
        \State $\mathbf{S}_{t+1} \gets$ allocated space 
        \While {aircraft remain}
            \For{each aircraft}
                \State $s_t \gets \mathbf{S}_{t}[aircraft]$
                \vspace{2pt}
                \State // {Build peaks from rewards in the environment}
                \vspace{2pt}
                \State $\mathbf{P^+} \gets $\textit{build pos rewards}
                \State $\mathbf{P^-} \gets $\textit{build neg rewards in Standard Positive Form}
                \State $\mathbf{P^*} \gets $\textit{build neg rewards for terrain in Standard Positive Form}
                \vspace{2pt}
                \State // Perform forward projection
                \vspace{2pt}
                \State $\mathbf{\Delta_1} \gets fwdProject(s_t, \mathbf{A}, \mathbf{L}, 0.1 ~s)$
                \State $\mathbf{\Delta_{10}} \gets fwdProject(s_t, \mathbf{A}, \mathbf{L}, 1.0 ~s)$
                \vspace{2pt}
                \State // Compute the value at each reachable state
                \vspace{2pt}
                \State $\mathbf{V^*} \gets $ \textit{allocate space for each reachable state}
                \For{$s_j \in \mathbf{\Delta_{10}}$}
                    \vspace{2pt}
                    \State // First for positive peaks 
                    \vspace{2pt}
                    \For{$p_i \in \mathbf{P^+}$}
                        \State // distance
                        \State $d_p \gets \norm{ s_j - \mathbf{location}(p_i) } _2 $ 
                        \State $r_p \gets \mathbf{reward}(p_i)$
                        \State $\gamma_p \gets \mathbf{discount}(p_i)$
                        \State $\mathbf{V^+}(p_i) \gets |r_p| \cdot \gamma_p ^ {d_p} $
                    \EndFor
                    \State $V_{max}^+ \gets \underset{p_i}{\max} ~ \mathbf{V^+} $
                    \vspace{2pt}
                    \State // Next for negative peaks (in Standard Positive Form) including terrain
                    \vspace{2pt}
                    \For{$n_i \in \{ \mathbf{P^-},\mathbf{P^*} \} $}
                        \State //distance
                        \State $d_n \gets \norm{ s_j - \mathbf{location}(n_i) } _2 $ 
                        \State $r_n \gets \mathbf{reward}(n_i)$
                        \State $\gamma_n \gets \mathbf{discount}(n_i)$
                        \State // within radius
                        \State $\rho_n \gets negDist_i < \mathbf{radius}(n_i) $ 
                        \State $\mathbf{V^-}(p_i) \gets int(\rho_n) \cdot |r_n| \cdot \gamma_n ^ {d_n} $
                    \EndFor
                    \State $V_{max}^- \gets \underset{p_i}{\max} ~\mathbf{V^-} $
                    \vspace{2pt}
                    \State // Hard deck penalty
                    \vspace{2pt}
                    \If{$\mathbf{altitude}(s_t) < penaltyAlt$}
                        \State $V_{deck} \gets 1000 - \mathbf{altitude}(s_t)$
                    \Else
                        \State $V_{deck} \gets 0$
                    \EndIf
                    \State $\mathbf{V^*}[s_t] \gets V_{max}^+ - V_{max}^- - V_{deck}$
                \EndFor
                
                \vspace{2pt}
                \State // Identify the most valuable action
                \vspace{2pt}
                \State $a^* \gets \underset{s}{\argmax}(\mathbf{V^*})$
                \vspace{2pt}
                \State // For illustration, the corresponding value
                \vspace{2pt}
                \State $maxValue \gets \mathbf{V^*}[ a^* ]$
                \vspace{2pt}
                \State // And the next state when taking the action
                \vspace{2pt}
                \State $\mathbf{s_{t+1}} \gets \mathbf{\Delta_1}[ a^* ]$
                
                \State $\mathbf{S}_{t+1}[ aircraft ] \gets \mathbf{s_{t+1}}$ 
            \EndFor
            
            \State // Now that all aircraft have selected an action, apply it
            \vspace{2pt}
            \State $\mathbf{S_t} \gets \mathbf{S}_{t+1}$
        \EndWhile
    \EndProcedure
    \end{algorithmic}
    \end{algorithm}

Peaks are described by the tuple $\mathbf{P}_i = \{ r_i, \gamma_i, \mathbf{\hat{p}}_i, R_i \}$ where the reward $r_i$ is the scalar value of the positive or negative reward scalar that is placed in the environment, the discount factor $\gamma_i \in ( 0, 1 )$ is used to determine discounted future reward, the position in 3D space is represented by $\mathbf{\hat{p}}_i$, and the radius of the peak is represented by $R_i$.  Positive peaks share the same discount factor of $\gamma_i$ which is also the discount factor for the overall MDP.  Negative rewards may have independent values of $\gamma_i$ and are not tied to the overall MDP's discount factor.  (See \cite{bertramMastersThesis} for a detailed explanation.)

The psuedocode for the FastMDP algorithm from \cite{bertram2020distributed} is shown in Algorithm \ref{orig_algo} for context in understanding the overall flow before the optimizations are described.  In summary, positive and negative peaks are constructed from the positive and negative rewards in the environment which are provided as inputs to the FastMDP algorithm.  Positive peaks are created from the positive rewards.  All negative rewards result in negative peaks which are constructed in \emph{standard positive form}, which means that negative rewards are treated temporarily by the algorithm as if they were positive, are used in generating the solution, and are reverted back to negative values before the algorithm generates its final answer.  \cite{bertramMastersThesis} describes this process in much more detail along with the theory behind it.

With the positive and negative peaks created, the algorithm then performs forward projection of the aircraft's state for each possible action that can be taken $a \in A$ over a planning window that (in this implementation) is 10 time steps into the future, resulting in a set of points $\mathbf{\Delta_{10}}$.  The value $\mathbf{V^*}[s_t]$ is computed at each point $s_t \in \mathbf{\Delta_{10}}$.  The action $a^*$ which leads to the most valuable action is then computed and is then selected for the next time step.
This process then causes the algorithm to take the optimal action that it can perform at each time step.

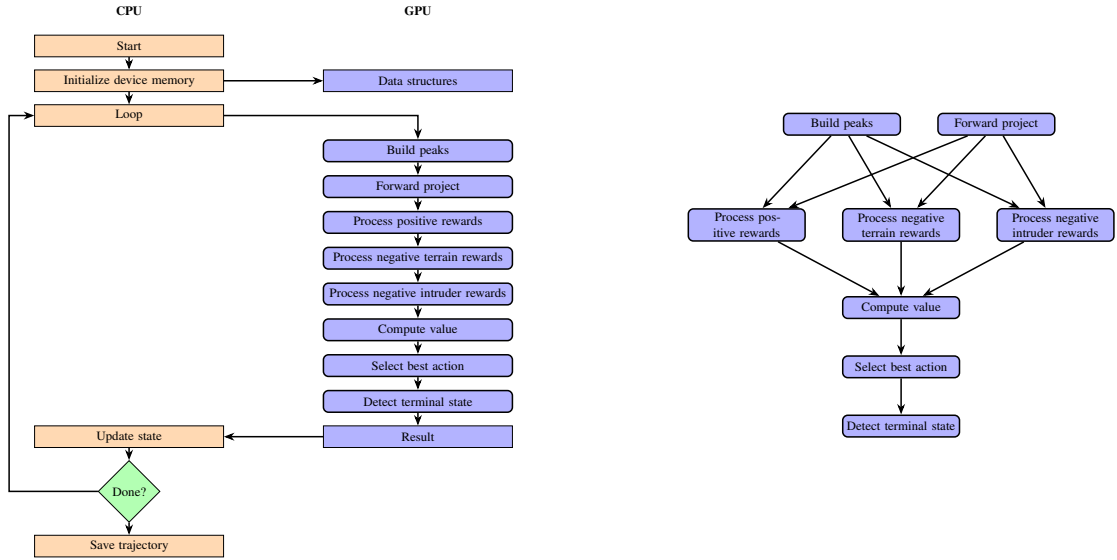
\begin{figure}[tbp]
\centering
\begin{subfigure}[t]{3in}
\begin{adjustbox}{scale={0.48}{0.48}, padding=0 0ex 0 0}
\begin{tikzpicture}[
      base/.style = {draw, text width=50mm, minimum width=50mm, minimum height=6mm,
                     align=center, 
                     },
 startstop/.style = {base, rectangle, rounded corners, fill=red!30},
   process/.style = {base, rectangle, fill=orange!30},
  cpuhdr/.style = {base, color=white, text=black, font=\bfseries },
  gpuhdr/.style = {cpuhdr, xshift=80mm},
  cpu/.style = {base, rectangle, fill=orange!30},
  buffer/.style = {base, rectangle, fill=green!30},
  gpu/.style = {base, fill=blue!30, xshift=80mm},
  kernel/.style = {gpu, rounded corners, very thick, fill=blue!30},
  decision/.style = {diamond, fill=green!30, text width=10mm, minimum width=10mm},
  ]
                    ]
\matrix[row sep=3.5mm, column sep=2mm] {
\node [cpuhdr]          {CPU}; 
\node [gpuhdr]          {GPU}; \\
\node [cpu]     (start) {Start}; \\
\node [cpu]     (init)  {Initialize device memory};
\node [gpu]     (data)  {Data structures}; \\
\node [cpu]     (loop)  {Loop}; \\
\node [kernel]  (a)     {Build peaks}; \\
\node [kernel]  (b)     {Forward project}; \\
\node [kernel]  (c)     {Process positive rewards}; \\
\node [kernel]  (d)     {Process negative terrain rewards}; \\
\node [kernel]  (e)     {Process negative intruder rewards}; \\
\node [kernel]  (f)     {Compute value}; \\
\node [kernel]  (g)     {Select best action}; \\
\node [kernel]  (h)     {Detect terminal state}; \\
\node [cpu]    (update) {Update state}; 
\node [gpu]    (result) {Result}; \\
\node [cpu, decision]    (done)   {Done?};  \\
\node [cpu]    (save)   {Save trajectory}; \\
};

\draw [arrows=-Stealth, very thick] 
     (done.west) -- 
     ([xshift=-70] done.west) -- 
     ([xshift=-7mm] loop.west) -- 
     (loop.west);
\draw [arrows=-Stealth, very thick] 
     (init) -- (data);

\draw [arrows=-Stealth, very thick] 
     (result) -- (update);

\draw [arrows=-Stealth, very thick]  (start) -- (init); 
\draw [arrows=-Stealth, very thick]  (init)  -- (loop); 

\draw [arrows=-Stealth, very thick]  (loop) -| (a); 
\draw [arrows=-Stealth, very thick]  (a) -- (b); 
\draw [arrows=-Stealth, very thick]  (b) -- (c); 
\draw [arrows=-Stealth, very thick]  (c) -- (d); 
\draw [arrows=-Stealth, very thick]  (d) -- (e); 
\draw [arrows=-Stealth, very thick]  (e) -- (f); 
\draw [arrows=-Stealth, very thick]  (f) -- (g); 
\draw [arrows=-Stealth, very thick]  (g) -- (h); 
\draw [arrows=-Stealth, very thick]  (h) -- (result); 

\draw [arrows=-Stealth, very thick]  (update) -- (done); 
\draw [arrows=-Stealth, very thick]  (done) -- (save); 
\end{tikzpicture}
\end{adjustbox}
\caption{Flowchart of the operations performed on the CPU versus the GPU.}
\label{myflowchart}
\end{subfigure}
\quad
\begin{subfigure}[t]{3in}
\centering
\begin{adjustbox}{scale={0.48}{0.48}, padding=0 0ex 0 0}
\begin{tikzpicture}[
    node distance = 20mm and 10mm,
      start chain = A going below,
      base/.style = {draw, text width=30mm, minimum width=30mm, minimum height=6mm,
                     align=center, 
                     },
  gpuhdr/.style = {cpuhdr, xshift=80mm},
  gpu/.style = {base, fill=blue!30},
  kernel/.style = {gpu, rounded corners, very thick, fill=blue!30},
  ]
                    
\node [kernel]  (a)  {Build peaks}; 
\node [kernel, right=of a]  (b)     {Forward project}; 
\node [kernel, below=of a.south east]  (d)     {Process negative terrain rewards}; 
\node [kernel, left =of d]  (c)     {Process positive rewards}; 
\node [kernel, right=of d]  (e)     {Process negative intruder rewards}; 
\node [kernel, below=15 mm of d]  (f)     {Compute value}; 
\node [kernel, below=10 mm of f]  (g)     {Select best action}; 
\node [kernel, below=10 mm of g]  (h)     {Detect terminal state}; 

\draw [arrows=-Stealth, very thick]  (a) -- (c); 
\draw [arrows=-Stealth, very thick]  (a) -- (d); 
\draw [arrows=-Stealth, very thick]  (a) -- (e); 
\draw [arrows=-Stealth, very thick]  (b) -- (c); 
\draw [arrows=-Stealth, very thick]  (b) -- (d); 
\draw [arrows=-Stealth, very thick]  (b) -- (e); 
\draw [arrows=-Stealth, very thick]  (c) -- (f); 
\draw [arrows=-Stealth, very thick]  (d) -- (f); 
\draw [arrows=-Stealth, very thick]  (e) -- (f); 
\draw [arrows=-Stealth, very thick]  (f) -- (g); 
\draw [arrows=-Stealth, very thick]  (g) -- (h); 

\node at (0,-12) {};  

\end{tikzpicture}
\end{adjustbox}
\caption{Data dependencies between the kernels.  Note that the kernels are scheduled serially by the CUDA library in the current implementation.}
\label{datadependencies}
\end{subfigure}
\caption{The algorithm is broken into multiple kernels which are scheduled in a pipeline on the GPU.}
\end{figure}

\section{Method}

The method used in this paper extends the FastMDP algorithm described in \cite{bertram2020distributed,bertram2020efficient} by reimplementing it for a Graphics Processing Unit (GPU) to take advantage of parallelism inherent in the algorithm which we term \emph{FastMDP-GPU}.  
The FastMDP algorithm implementations previously published 
 in \cite{bertram2020distributed,bertram2020efficient,bertramMastersThesis} are efficient in that they achieve $O(n)$ performance where $n$ is the number of positive and negative rewards.  However, as the number of rewards scales to large numbers, even this $O(n)$ scaling is not enough to achieve real-time performance above 50-100 aircraft, depending on the particular problem.  
  
In contrast, the GPU based approach allows many parts of the problem to be computed in parallel, resulting in an overall reduction in computation time required for the same set of inputs.
We use this GPU based method to study the level of scaling that can be achieved by the algorithm by applying it to a pre-departure flight planning problem where conflict free flight path trajectories are computed in a first-come-first-served order.

The overall FastMDP-GPU algorithm used in this paper is shown in Figure \ref{myflowchart}.  The Central Processing Unit (CPU) code is responsible for initialization and coordination of the processing flow, and the GPU side is responsible for performing the parallel computations.  The GPU code is implemented using NVIDIA's CUDA library and the GPU code blocks are referred to as \emph{kernels}.  Each kernel is designed to be a simple block of code with as few loops and branches as possible.  When a kernel is launched by the CPU, on the GPU side many instances of the code within the kernel are launched in parallel \emph{threads}.  The threads are scheduled by the CUDA library to run on the available GPU cores in the hardware and run in batches until all threads have completed.  
The GPU is designed to handle large numbers of threads with very little overhead for switching between threads (unlike CPUs).
It is not unusual for a kernel launch to run millions of threads on the GPU.

In CUDA, threads are organized into blocks, which are in turn organized into a grid.  The blocks within a grid can be indexed by 1D, 2D, or 3D indices.  Likewise, the threads within a block can also be indexed by 1D, 2D, or 3D indices. 
In the implementation used in this paper, the kernels shown in Figure \ref{myflowchart} use different indexing schemes as needed in order to maximize parallelism and are summarized in Table \ref{kernel_indexing}.  The kernels are arranged in a pipeline with the sequence defined by the CPU.  The kernels themselves run serially, but the threads of each kernel run in parallel with each other.  Individual thread scheduling is managed by the CUDA library and GPU.  

\begin{table}[tbp]
\caption{Major data structures in GPU memory as inputs or outputs of each kernel, where $N$ indicates the number of aircraft being simulated in the batch, $A$ indicates the number of actions that can be taken, $R_p$ indicates the number of positive rewards, $R_t$ indicates the number of negative terrain rewards, $R_i$ indicates the number of negative intruder rewards, and $W$ indicates the time step window of $10$.  Data structures which are multi-field structures are indicated including the number of fields in the dimensionality (e.g., $N \times 6$ for a structure with 6 fields.)  Sample values shown for $N=1, A=1350, R_p=1, R_t=50,R_i=2000$ which are typical values used for a batch size of 1, an action space resulting in $1350$ possible actions at each time step, terrain modelled with $50$ negative rewards, and $2000$ intruders.}
\label{kernel_indexing}
\begin{center}
\rowcolors{1}{LtGray}{}
\begin{tabular}{ |c|p{125pt}|c|c|p{100pt}| } 
 \hline
 \rowcolor{Gray}
 \textbf{Symbol} & \textbf{Purpose} & \textbf{Data Type} & \textbf{Dimensionality}  & \textbf{Sample size (bytes)} \\
 \hline
 $\mathbf{P^+}$ & Peaks formed from positive rewards (eg., the goal) & 64-bit float & $R_p \times 6$ & $1 \times 6 \times 8 = 48$ \\
 $\mathbf{P^-}$ & Peaks formed from negative rewards from other aircraft in the batch & 64-bit float & $5 \times (N-1) \times 6$ & $5 \times 0 \times 6 \times 8 = 0$ \\
 $\mathbf{P^I}$ & Peaks formed from negative rewards from intruders & 64-bit float & $ 5 \times R_i \times 6$ & $5 \times 2000 \times 6 \times 8 = 480,000$ \\
 $\mathbf{P^T}$ & Peaks formed from negative rewards from terrain & 64-bit float & $R_t \times 6  $ & $50 \times 6 \times 8 = 2,400$ \\
 $\mathbf{\Delta_{10}}$ & States resulting from forward projection & 64-bit float & $N \times A \times W \times 12$ & $ 1 \times 1350 \times 10 \times 12 \times 8 = 1,296,000 $  \\
 $\mathbf{V^+}$ & Value contributed at states due to contributions from positive rewards from $\mathbf{P^+}$ & 64-bit float & $N \times A \times W$ & $ 1 \times 1350 \times 10 \times 8 = 108,000$ \\
 $\mathbf{V^-}$ & Value contributed at states due to contributions from negative rewards from $\mathbf{P^-}$ & 64-bit float & $N \times A \times W$ & $  108,000$ \\
 $\mathbf{V^I}$ & Value contributed at states due to contributions from negative rewards from $\mathbf{P^I}$ & 64-bit float & $N \times A \times W$ & $108,000$ \\
 $\mathbf{V^T}$ & Value contributed at states due to contributions from negative rewards from $\mathbf{P^T}$ & 64-bit float & $N \times A \times W$ & $108,000$ \\
 $\mathbf{V}$ & Value computed from all positive and negative contributions (primarily for debug and visualization)  & 64-bit float & $N \times A \times W$ & $108,000$ \\
 $\mathbf{V^*}$ & Value computed from all positive and negative contributions & 64-bit float & $N \times W$  & $1 \times 10 \times 8 = 80$ \\
 $\mathbf{A^*}$ & Selected action for each aircraft & 64-bit float & $N$  & $1 \times 8 = 8$ \\
\hline
\end{tabular}
\end{center}
\end{table}

Each kernel consumes one or more data structures defined in the GPU card's memory and outputs results into one or more other data structures in the GPU memory.  Copying memory between the CPU and GPU is minimized and is primarily performed during initialization to set up the GPU state before the algorithm runs.  At the end of each cycle, a minimal amount of state information is copied from the GPU memory to the CPU memory so that the CPU software is aware of the current state of the simulation.  The major data structures used as inputs and outputs of kernels are described in Table \ref{data_structures}.

\begin{table}[tbp]
\caption{Kernel indexing schemes used for each kernel, where $N$ indicates the number of aircraft being simulated in the batch, $A$ indicates the number of actions that can be taken, $R_p$ indicates the number of positive rewards, $R_t$ indicates the number of negative terrain rewards, and $R_i$ indicates the number of negative intruder rewards.  Sample values shown for $N=1, A=1350, R_p=1, R_t=50,R_i=2000$ which are typical values used for a batch size of 1, an action space resulting in $1350$ possible actions at each time step, terrain modelled with $50$ negative rewards, and $2000$ intruders.}
\label{data_structures}
\begin{center}
\rowcolors{1}{LtGray}{}
\begin{tabular}{ |c|c|c|c|c|c|c|c|c|c| } 
 \hline
 \textbf{Name} & \textbf{Dimensionality} & \textbf{Indexes} & \textbf{Sample number of threads} \\
 \hline
 Build peaks & 1D & $N$ & 1 \\
 Forward project & 2D & $N \times A$ & 1350 \\
 Process positive rewards & 3D & $N \times A \times R_p$ & 1350 \\
 Process negative rewards & 3D & $5 \times N \times A \times (N-1)$ & 0 \\
 Process negative terrain rewards & 3D & $N \times A \times R_t$ & 67,500 \\
 Process negative intruder rewards & 3D & $5 \times N \times A \times R_i$ & 13,500,000 \\
 Compute value & 2D & $N \times A$ & 1350 \\
 Select best action & 1D & $N$ & 1 \\
 Determine terminal state & 3D & $N \times N $ & 1 \\
 \hline
\end{tabular}
\end{center}
\end{table}

Each kernel is now described in detail.

\subsection{Kernel: Build peaks}

Peaks are built as described in Table \ref{reward_table}.  
In this implementation, each aircraft is assigned a single goal location which represents a vertiport or other landing site.  A single positive peak is created to model the goal.
For each intruder, multiple risk wells (negative rewards) are defined at different points along the intruder's current trajectory as defined by the position and the linear velocity of the intruder.  

The inputs of the algorithm are the current state of the aircraft in the batch and the intruders.  The outputs of the kernel are the positive peaks $\mathbf{P^+}$, negative peaks $\mathbf{P^-}$, and negative peaks for intruders $\mathbf{P^I}$.  

Note that terrain features are also modelled with risk wells, but these peaks $\mathbf{P^T}$ are defined statically at load time and transferred to GPU memory during the algorithm initialization phase.  Also note that this kernel is called once for each aircraft in the batch (see Table \ref{kernel_indexing}).  While it contains loops, profiling has shown that the kernel contributes negligible overhead and needs no further optimization.  The logic in this kernel is more suitable for operation on CPU, but it is implemented as a kernel primarily to avoid unnecessary copying to and from CPU and GPU memory.

\begin{algorithm}[H]
    \footnotesize
    \caption{Build Peaks Kernel} 
    \begin{algorithmic}[1]
    \Procedure{Build Peaks}{$i_{ac}$}
        \State // $i_{ac}$: Aircraft index
        \State // $i_{a}$: Action index
        \vspace{10pt}
        \State // {Build peaks per Table \ref{reward_table}}
        \vspace{2pt}
        \State // {Build positive for the goal}
        \State $\mathbf{P^+} \gets \{ r_g, \gamma = 0.9, \mathbf{\hat{p}}_g, \infty \}  $  // goal
        \State // {Build negative rewards for each other aircraft in the batch}
        \For{$ac \in batch$}
            \State $\mathbf{P^-}_{ac} \gets \{ r_{ac}, \gamma_{ac}, \mathbf{\hat{p}}_{ac}, R_{ac} \}  $  // aircraft
        \EndFor
        \State // {Build negative rewards for each intruder}
        \For{$ac \in intruders$}
            \State $\mathbf{P^I}_{ac} \gets \{ r_{ac}, \gamma_{ac}, \mathbf{\hat{p}}_{ac}, R_{ac} \}  $  // aircraft
        \EndFor
    \EndProcedure
    \end{algorithmic}
    \end{algorithm}

\begin{table*}[tbp]
\vspace{5pt}
\caption{Peaks created in the environment. For each aircraft in the environment, multiple negative peaks are placed along its trajectory.  Terrain features in this implementation are negative rewards placed manually to overlay the terrain.  Each aircraft's goal is also manually selected and represents a vertiport in the environment.  Here $\mathbf{\hat{p}}$ represents the position of an aircraft and $\mathbf{\hat{v}}$ represents the velocity of the aircraft.}
\label{reward_table}
\begin{center}
\begin{tabular}{ |c|c|c|c|c|l| } 
 \multicolumn{6}{l}{\textit{For each intruder or other aircraft in the batch:}} 
 \vspace{4pt}
 \\
 \hline
 \rowcolor{Gray}
 Magnitude  & Decay factor & Location & Radius & Timesteps & Comment \\
 \hline
 $-1000$ & $.97$ & $\mathbf{\hat{p}} + \mathbf{\hat{v}} t $ & $300 + 10 t$ & $\forall t \in \{ -5, 0, 5, 10, 15\}$  & Collision avoidance, 5 rewards \\
 \hline
 
 \multicolumn{6}{l}{} \\
 \multicolumn{6}{l}{\textit{For each terrain feature:}} 
 \vspace{4pt}
 \\
 \hline
 \rowcolor{Gray}
 Magnitude  & Decay factor & Location & Radius & Timesteps & Comment \\
 \hline
 $-1000$ & $.99$ & manually placed & manually selected & N/A  & Terrain avoidance \\
 \hline
 
 \multicolumn{6}{l}{} \\
 \multicolumn{6}{l}{\textit{For aircraft's goal:}} 
 \vspace{4pt}
 \\
 \hline
 \rowcolor{Gray}
 Magnitude  & Decay factor & Location & Radius & Timesteps & Comment \\
 \hline
 $200$ & $.999$ & manually placed & $\infty$  & N/A  & Vertiport attraction \\
 \hline
%
%
\end{tabular}
\end{center}
\vspace{-15pt}
\end{table*}

\subsection{Kernel: Forward project}

Forward projection here refers to using models of the aircraft dynamics to compute the future state of the aircraft based on an assumed action for a fixed duration of time.  The aircraft dynamics and actions used in this paper are the same as those used in \cite{bertram2020distributed}.  The forward projection used here is considered a module that can be replaced with another physics model or forward projection method.  Additionally, multiple physics models that model different aircraft types could also be implemented and used to simulate different aircraft dynamics within the same simulation.

The input of this kernel is the current state of all aircraft in the batch and the set of all possible actions that an aircraft can take from the current state.  One thread is created for each aircraft in the batch and each action that can be taken from the current state for $N \times A$ total threads (see Table \ref{kernel_indexing}).

The output of this kernel is the future state of all aircraft in the batch for each possible action for each time step in the lookahead window.

The time window forward projection is performed over is denoted with $W$ and represents the number of simulation time steps of duration $dt$ to perform.  In this implementation, $W=10$ and each time step is $dt=0.1$ seconds.  The time window could be increased or decreased and is selected so that the forward projected actions provide a significant enough spread in the state space for the algorithm to detect a difference in value between states.  The key here is that the forward projection needs to be far enough away for the agent to react to the truncated boundary of risk wells.  Less maneuverable aircraft models will require a larger forward projection window $W$.

Note here that nothing precludes the time step $dt$ from being variable.  For simplicity, in this implementation the time step is fixed, but if an adaptive time step $dt$ or time window $W$ were desired, this is achievable without loss of generality.  Likewise, at different time scales, different dynamics models could be used which are appropriate for the timescale if an adaptive fidelity approach were desired.

Note also that if the dynamics model is such that the points in time can be computed independently from each other, then the kernel could be further parallelized, but we do not assume this to always be the case.

\begin{algorithm}[H]
    \footnotesize
    \caption{Forward Projection Kernel} 
    \begin{algorithmic}[1]
    \Procedure{Forward Projection}{$i_{ac}, i_{a}$}
        \State // $i_{ac}$: Aircraft index
        \State // $i_{a}$: Action index
        \vspace{10pt}
        \State // Perform forward projection of aircraft dynamics given action
        \vspace{2pt}
        \For{$t \in \{1,\cdots,W\}$}
            \State $\mathbf{\Delta_{10}}[i_{ac},i_{a},t] \gets $ step physics computations forward in time
        \EndFor
    \EndProcedure
    \end{algorithmic}
    \end{algorithm}
\subsection{Kernel: Process positive rewards}

For each of the future states computed by forward projection, the value contributed by positive peaks is computed and saved in an output array.

The input of this kernel is the forward projected state for each action for each time step $dt$ in the forward projection window $W$.  One thread is created for each aircraft in the batch for each action that can be taken from the current state for each positive peak for $N \times A \times R_{p}$ total threads (see Table \ref{kernel_indexing}).

The output of this kernel is the value at each state that is contributed by each positive peak.

\begin{algorithm}[H]
    \footnotesize
    \caption{Positive Rewards Kernel} 
    \begin{algorithmic}[1]
    \Procedure{Positive Rewards}{$i_{ac}, i_{a}, i_{p}$}
        \State // $i_{ac}$: Aircraft index
        \State // $i_{a}$: Action index
        \State // $i_{p}$: Peak index
        \vspace{10pt}
        \For{$t \in \{1,\cdots,W\}$}
                \State // Get projected state for this time step
                \State $s \gets \mathbf{\Delta_{10}}[ i_{ac}, i_{a}, t]$
                \State // Get peak
                \State $p_i \gets \mathbf{P^+}[i_{p}]$
        
                \State // Compute distance between state and peak
                \State $d_p \gets \norm{ s - \mathbf{location}(p_i) } _2 $ 
                \State // Extract reward magnitude for the peak from data structure
                \State $r_p \gets \mathbf{reward}(p_i)$
                \State // Extract discount factor for the peak from data structure
                \State $\gamma_p \gets \mathbf{discount}(p_i)$
                \State // Compute the value with respect to the peak
                \State $V \gets |r_p| \cdot \gamma_p ^ {d_p} $
                \State // Save max value (atomic operation)
                \State $\mathbf{V^+}[i_{ac},i_{a},t] \gets \max \left(\mathbf{V^+}[i_{ac},i_{a},t], V \right)$
        \EndFor
    \EndProcedure
    \end{algorithmic}
    \end{algorithm}

\subsection{Kernel: Process negative rewards}

For each of the future states computed by forward projection, the value contributed by negative peaks is computed and saved in an output array.

The input of this kernel is the forward projected state for each action for each time step $dt$ in the forward projection window $W$.  One thread is created for each aircraft in the batch for each action that can be taken from the current state for each negative peak for $N \times A \times R_{n}$ total threads (see Table \ref{kernel_indexing}).

The output of this kernel is the value at each state that is contributed by each negative peak.

\begin{algorithm}[H]
    \footnotesize
    \caption{Negative Rewards Kernel} 
    \begin{algorithmic}[1]
    \Procedure{Negative Rewards}{$i_{ac}, i_{a}, i_{p}$}
        \State // $i_{ac}$: Aircraft index
        \State // $i_{a}$: Action index
        \State // $i_{p}$: Peak index
        \vspace{10pt}
        \For{$t \in \{1,\cdots,W\}$}
                \State // Get projected state for this time step
                \State $s \gets \mathbf{\Delta_{10}}[ i_{ac}, i_{a}, t]$
                \State // Get peak
                \State $p_i \gets \mathbf{P^-}[i_{p}]$
        
                \State // Compute distance between state and peak
                \State $d \gets \norm{ s - \mathbf{location}(p_i) } _2 $ 
                \State // Extract reward magnitude for the peak from data structure
                \State $r \gets \mathbf{reward}(p_i)$
                \State // Extract discount factor for the peak from data structure
                \State $\gamma \gets \mathbf{discount}(p_i)$
                \State // Extract radius for the peak from data structure
                \State $R \gets \mathbf{radius}(p_i)$
                \State // Compute whether we are inside the radius of the peak, result is a 1 if true or a 0 if false.
                \State $in \gets d < R$
                \State // Compute the value with respect to the peak
                \State $V \gets in \cdot |r| \cdot \gamma ^ {d} $

                \State // Save max value (atomic operation)
                \State $\mathbf{V^-}[i_{ac},i_{a},t] \gets \max \left( \mathbf{V^-}[i_{ac},i_{a},t], V \right)$
        \EndFor
    \EndProcedure
    \end{algorithmic}
    \end{algorithm}
    
\subsection{Kernel: Process negative terrain rewards}

For each of the future states computed by forward projection, the value contributed by terrain peaks is computed and saved in an output array.

The input of this kernel is the forward projected state for each action for each time step $dt$ in the forward projection window $W$.  One thread is created for each aircraft in the batch for each action that can be taken from the current state for each terrain peak for $N \times A \times R_{t}$ total threads (see Table \ref{kernel_indexing}).

The output of this kernel is the value at each state that is contributed by each terrain peak.

\begin{algorithm}[H]
    \footnotesize
    \caption{Terrain Rewards Kernel} 
    \begin{algorithmic}[1]
    \Procedure{Terrain Rewards}{$i_{ac}, i_{a}, i_{p}$}
        \State // $i_{ac}$: Aircraft index
        \State // $i_{a}$: Action index
        \State // $i_{p}$: Peak index
        \vspace{10pt}
        \For{$t \in \{1,\cdots,W\}$}
                \State // Get projected state for this time step
                \State $s \gets \mathbf{\Delta_{10}}[ i_{ac}, i_{a}, t]$
                \State // Get peak
                \State $p_i \gets \mathbf{P^T}[i_{p}]$

                \State // Compute distance between state and peak
                \State $d \gets \norm{ s - \mathbf{location}(p_i) } _2 $ 
                \State // Extract reward magnitude for the peak from data structure
                \State $r \gets \mathbf{reward}(p_i)$
                \State // Extract discount factor for the peak from data structure
                \State $\gamma \gets \mathbf{discount}(p_i)$
                \State // Extract radius for the peak from data structure
                \State $R \gets \mathbf{radius}(p_i)$
                \State // Compute whether we are inside the radius of the peak, result is a 1 if true or a 0 if false.
                \State $in \gets d < R$
                \State // Compute the value with respect to the peak
                \State $V \gets in \cdot |r| \cdot \gamma ^ {d} $

                \State // Save max value (atomic operation)
                \State $\mathbf{V^T}[i_{ac},i_{a},t] \gets \max \left(\mathbf{V^T}[i_{ac},i_{a},t], V \right)$
        \EndFor
    \EndProcedure
    \end{algorithmic}
    \end{algorithm}
    
\subsection{Kernel: Process negative intruder rewards}

For each of the future states computed by forward projection, the value contributed by intruder peaks is computed and saved in an output array.

The input of this kernel is the forward projected state for each action for each time step $dt$ in the forward projection window $W$.  
One thread is created for each aircraft in the batch for each action that can be taken from the current state for each intruder peak for $N \times A \times R_{i}$ total threads (see Table \ref{kernel_indexing}).

The output of this kernel is the value at each state that is contributed by each intruder peak.

\begin{algorithm}[H]
    \footnotesize
    \caption{Intruder Rewards Kernel} 
    \begin{algorithmic}[1]
    \Procedure{Intruder Rewards}{$i_{ac}, i_{a}, i_{p}$}
        \State // $i_{ac}$: Aircraft index
        \State // $i_{a}$: Action index
        \State // $i_{p}$: Peak index
        \vspace{10pt}
        \For{$t \in \{1,\cdots,W\}$}
                \State // Get projected state for this time step
                \State $s \gets \mathbf{\Delta_{10}}[ i_{ac}, i_{a}, t]$
                \State // Get peak
                \State $p_i \gets \mathbf{P^I}[i_{p}]$

                \State // Compute distance between state and peak
                \State $d \gets \norm{ s - \mathbf{location}(p_i) } _2 $ 
                \State // Extract reward magnitude for the peak from data structure
                \State $r \gets \mathbf{reward}(p_i)$
                \State // Extract discount factor for the peak from data structure
                \State $\gamma \gets \mathbf{discount}(p_i)$
                \State // Extract radius for the peak from data structure
                \State $R \gets \mathbf{radius}(p_i)$
                \State // Compute whether we are inside the radius of the peak, result is a 1 if true or a 0 if false.
                \State $in \gets d < R$
                \State // Compute the value with respect to the peak
                \State $V \gets in \cdot |r| \cdot \gamma ^ {d} $

                \State // Save max value (atomic operation)
                \State $\mathbf{V^I}[i_{ac},i_{a},t] \gets \max \left(\mathbf{V^I}[i_{ac},i_{a},t], V \right)$
        \EndFor
    \EndProcedure
    \end{algorithmic}
    \end{algorithm}
    
\subsection{Kernel: Compute value}

In this pipeline stage, the results of all positive and all negative rewards are combined into a single value for each state computed by forward projection.  
One thread is created for each aircraft in the batch for each action that can be taken from the current state for $N \times A $ total threads (see Table \ref{kernel_indexing}).

\begin{algorithm}[H]
    \footnotesize
    \caption{Compute Value Kernel}
    \begin{algorithmic}[1]
    \Procedure{Compute Value}{$i_{ac}, i_{a}$}
        \State // $i_{ac}$: Aircraft index
        \State // $i_{a}$: Action index
        \vspace{10pt}
        \State // Initialize the total value
        \State $V_{max} \gets 0$
        \State // Compute the maximum value that this action resulted in along its trajectory
        \For{$t \in \{1,\cdots,W\}$}
            \State // Get the maximum positve, negative, terrain, and intruder values at this time step.
            \State $V^+_{max} \gets \mathbf{V^+}[i_{ac},i_{a},t]$ 
            \State $V^-_{max} \gets \mathbf{V^-}[i_{ac},i_{a},t]$ 
            \State $V^-_{max} \gets \mathbf{V^T}[i_{ac},i_{a},t]$ 
            \State $V^T_{max} \gets \mathbf{V^I}[i_{ac},i_{a},t]$ 
           
            \State // Compute an altitude penalty if we go below a hard deck minimum altitude 
            \State $V_{alt} \gets $ apply penalty if less than hard deck
            
            \State // Compute the value from all the components
            \State $V \gets V^+_{max} - \max \left( V^-_{max}, V^T_{max}, V^I_{max} \right) - V_{alt}$
            \State $V_{max} \gets \max \left( V_{max}, V \right)$
            
        \EndFor
        \State // Save off the maximum value this this action achieved
        \State $\mathbf{V^*}[i_{ac},i_{a}] \gets V_{max}$
    \EndProcedure
    \end{algorithmic}
    \end{algorithm}
\subsection{Kernel: Select best action}

Here the value of each possible action is computed and the action with maximum value is identified and selected.
One thread is created for each aircraft in the batch for $N$ total threads (see Table \ref{kernel_indexing}).

\begin{algorithm}[H]
    \footnotesize
    \caption{Select Action Kernel}
    \begin{algorithmic}[1]
    \Procedure{Select Action}{$i_{ac}$}
        \State // $i_{ac}$: Aircraft index
        \vspace{10pt}
        \State // Identify the most valuable action
        \State $a^* \gets \max_{a \in A} \mathbf{V^*}[i_{ac},a]$
        \State $\mathbf{A^*}[i_{ac}] = a^*$
    \EndProcedure
    \end{algorithmic}
    \end{algorithm}
    
\subsection{Kernel: Determine terminal state}

This function monitors all aircraft in the simulation to detect collisions, Near Mid-Air Collisions (NMACs), collisions with terrain, and aircraft that have successfully reached their goals.  Selected actions are also applied to the internal simulation state which effectively advances simulation time by $dt$.  This code operates on the GPU side to avoid having to transfer memory to the CPU.  A summary of the results of this function and of the current simulation state are passed to the CPU side to determine if simulation should continue or end.  As this is just simple accounting and collision testing, the code is omitted.

\section{Experimental Setup}

To demonstrate the effectiveness of the algorithm in solving a real-world problem, we apply the algorithm to a pre-departure flight planning (PDFP) problem in which an aircraft must determine a flight plan before takeoff which has no conflicts with any other aircraft's previously accepted flight plan.  In this setup, an aircraft submits a starting and ending location to the PDFP system, and the PDFP system returns a trajectory to the aircraft which is known to be conflict free.  The PDFP system maintains a database of terrain and of \emph{accepted} flight plans, which are the result of previous successful requests for flight plans from other aircraft who have utilized the first-come-first-served PDFP request system.  The PDFP system guarantees that the result of a PDFP request is conflict free with respect to terrain and with all accepted flight plans.  When a new flight plan is generated by the PDFP system, it is automatically stored in the database of accepted flight plans and is then used for any new future requests.  

This could represent a centralized planning system where all aircraft submit to a central service provider, such as NASA's UTM.  This could also represent a distributed case where an operator of a fleet or individual aircraft build a flight plan using a published set of flight plans.  In this case, multiple operators could build flight plans simultaneously that are conflict free with respect to the published set of flight plans.  However, there would need to be some mechanism which resolves conflicting flight plans that result from distributed planners who inadvertently plans which conflict with each other but are conflict free with respect to the published flight plans.  Such a mechanism is outside the scope of this paper.

The implementation for this paper is focused on determining the level of scalability that can be achieved by the algorithm and does not put any effort into the database design.  The database used here is a simple table in memory which stores a list of previously accepted conflict free flight plans.  This database is loaded from a file and a mechanism is available to add an accepted flight plan to this file.  In a larger scale implementation for a production environment, this could be implemented by a production quality database system such as SQL server or equivalent.
Likewise, in this implementation, it is naively assumed that all simulation time steps should take place at a $0.1$ second increment.  In a more sophisticated implementation, a larger time step could be taken in regions where it is safe to do so (e.g., away from other aircraft and terrain).  This would have the effect of being able to complete a flight plan in many fewer iterations of the algorithm and would lead to a higher throughput of the pre-departure flight planning system, but would not impact the level of scalability of the algorithm itself being studied in this paper.

While the implementation in this paper plans for a single source and destination, it can easily be extended to plan along a route using known points (e.g., NAVAIDs) as waypoints.  In this way, multiple routes could be explored in parallel by allowing the algorithm to generate several candidate schedules which are returned to the requester.  The requester could then evaluate different attributes of the flight plan, such as fuel usage, risk profile, etc and select a preferred route.  Likewise by performing parallel scheduling at different departure times, varying routes can be examined.  For example, it may be that leaving 30 minutes earlier avoids known traffic congestion and leads to lower fuel cost.

Each request causes the algorithm in this paper to be invoked with one or more aircraft.  All previously accepted flight plans are treated as intruders and are made available to the algorithm.  The algorithm is allowed to execute until it reaches its goal or a collision occurs during simulation.  Collisions with terrain or with other aircraft are tracked and reported.  Only a trajectory without collisions is accepted, otherwise an error is reported.

For batch sizes greater than one, there are two ways in which to operate this algorithm.  One possibility is to invoke the algorithm with multiple aircraft being co-simulated together with the same set of accepted flights.  In this mode, the aircraft will be aware of each other and will route around each other and all accepted flight plans.  This is necessary if the aircraft within the batch might possibly intersect each other's flight plans (e.g., when they are flying through an overlapping flight volume.)  In cases where requests can be segregated into non-overlapping flight volumes (e.g. separate sections of a metropolitan area, separate sectors of airspace, etc), then independent parallel instances of the algorithm can be run independent of each other.  For a successful large scale implementation of this algorithm, a strategy should be employed to break requests into manageable numbers of overlapping flights, and to co-simulate small numbers of flights together in batches allowing independent batches to run in parallel.  An implementation of this level of parallelism is left to future work.

While in this paper, all accepted flight plans are generated by successively running the algorithm with a sequence of source and destinations, in principle the flight plans could also be generated by some external source and imported into this algorithm.  The flight plans could be in the form of trajectory points as is currently done in this implementation, or could be in some higher level summary form such as line segments, Bezier splines, or some other more efficient representation.

\section{Preliminary Results}

\begin{figure}[tbp]
\centering
\begin{subfigure}[t]{3in}
\begin{adjustbox}{scale=.75, padding=0 0ex 0 0}
\begin{tikzpicture}
\begin{axis}[
    title={Performance versus number of accepted flight plans},
    xlabel={Number of accepted flight plans},
    ylabel={Frames per second [Hz] },
    xmin=0, xmax=3000,
    ymin=0, ymax=200,
    xtick={0,1000,2000,3000},
    ytick={0,10,50,100,150,200},
    legend pos=north east,
    ymajorgrids=true,
    grid style=dashed,
]

\addplot[
    color=blue,
    mark=square,
    ]
    coordinates {
    (1,187.6)(125,85.0)(250,52.0)(500,29.9)(1000,16.2)(2000,8.6)(3000,5.69)
    };
    
\addplot[
    color=orange,
    mark=triangle,
    ]
    coordinates {
    (1,423.7)(125,166.0)(250,93.4)(500,50.0)(1000,25.9)(2000,13.2)(3000,8.89)
    };

\addplot[
    color=green,
    mark=triangle,
    ]
    coordinates {
    (3,28.6)(15,22.7)(30,17.2)(45,14.83)(60,12.9)(75,10.5)(90,8.58)(150,6.15)
    };
    \legend{GeForce 2080 RTX, GeForce Titan Xp, CPU version}
\end{axis}
\end{tikzpicture}
\end{adjustbox}
\caption{Performance results as the number of accepted flight plans varies.  GPU performance greatly exceeds the CPU performance from \cite{bertram2020distributed}}
\label{perf1}
\end{subfigure}
\quad
\begin{subfigure}[t]{3in}
\centering
\begin{adjustbox}{scale=.75, padding=0 0ex 0 0}
\begin{tikzpicture}
\begin{axis}[
    title={Performance versus batch size for fixed number $100$ of intruders},
    xlabel={Batch size},
    ylabel={Frames per second (fps) [Hz] },
    xmin=0, xmax=20,
    ymin=0, ymax=350,
    xtick={0,5,10,15,20},
    ytick={0,10,50,100,150,200,250,300},
    legend pos=north east,
    ymajorgrids=true,
    grid style=dashed,
]

\addplot[
    color=red,
    mark=triangle,
    ]
    coordinates {
    (1,194.5)(5,48.5)(10,24.0)(20,12.2)
    }; \label{plot_one};

\addplot[
    color=green,
    mark=square,
    ]
    coordinates {
    (1,194.5)(5,242.5)(10,240.0)(20,244.2)
    }; \label{plot_two};

    \legend{Batch cycles completed,Total cycles completed}
\end{axis}

\end{tikzpicture}
\end{adjustbox}
\caption{Performance results as the batch size varies.}
\label{perf2}
\end{subfigure}
\caption{Performance results of the algorithm.}
\end{figure}
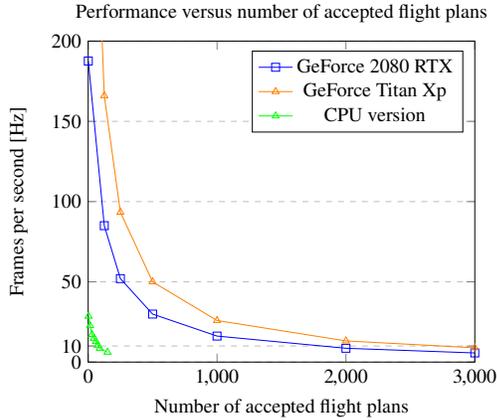
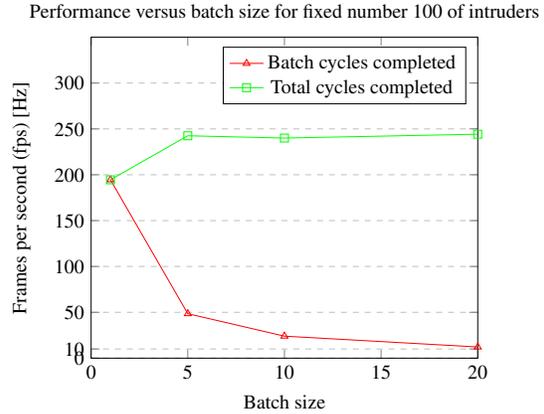

Figure \ref{perf1} shows results for the algorithm's performance as the number of accepted flight plans increases.  The GeForce 2080 RTX used for this test has 8GB of RAM and tests were performed with NVIDA driver version 441.66 running CUDA 10.2 on Windows 10.  The GeForce Titan XP has 12GB of RAM, used driver 440.60 running CUDA 10.2 on Ubuntu 18.04.



Performance results as as the number of accepted flight plans varies are shown in Figure \ref{perf1}.  Any value above 10 Hz represents running faster than real-time.  This crossover occurs near 2,000 accepted flight plans on the GeForce 2080 RTX, which represents approximately 10,000 peaks that are processed by the algorithm.  On the GeForce Titan XP, the crossover occurs at about 3,000 accepted flight plans.  On the CPU version, this crossover happens around 75 flight plans.

Performance results as the batch size varies are shown in Figure \ref{perf2}.  While this demonstrates that the frame rate for processing the batch decreases as the number of aircraft in the batch increase, it also shows that if the total cycles processed (the $x$-coordinate multiplied by the $y$-coordinate) is considered, it reflects that the GPU becomes saturated and can perform a certain amount of work (certain number of computations) per second.  Once the GPU becomes saturated (at $x=5$ on the graph), it can no longer keep up with the amount of new work being requested.  This saturation point will occur at a different level depending on the power of the GPU.

\section{Expected Outcome}

\textit{Note to reviewers:  For this extended abstract submission, we have implemented the core algorithm and have proven the computational viability and scalability of the algorithm.  For the final paper we plan on evaluating the performance of the algorithm along a similar line of exploration as \cite{guerreiro2019mission} and specifically want to generate candidate flight plans with different ground delay to allow the operator to select among viable flight plans.  We would also like to perform measurements that show replanning events due to unplanned pop-events occurring, such as a weather system moving through an area or a TFR due to an emergency (air based) vehicle needing to reserve airspace.  We believe our algorithm should be able to efficiently replan in these cases and we would like to explore this possibility.  If we can obtain the same data used in one of the cited papers, we would like to compare our results with these published results.  If the data is not available, we plan on defining a set of benchmarks with enough detail that future researchers will be able to use them to compare their results to ours.  We also hope to be able to release our generated flight plans as a reference set of flight plans that others might use for their own research.}

\section{Conclusion}

This paper presents FastMDP-GPU, an approach for performing pre-departure flight planning that is efficient and scales to a large number of aircraft using a highly parallelized GPU-based approach.

For future work, further optimization can be performed on the algorithm to obtain higher levels of utilization of the GPU.  Additionally, multi-GPU systems can be utilized to perform concurrent processing on multiple GPUs.  Likewise, moving processing into the cloud or a cluster so that multiple GPUs on separate systems are utilized could also improve performance.  

Given the limitations of a particular GPU, another strategy used would be to segment the problem by geographical area.  Geographic Information Systems (GIS) databases often employ queries which operate over an area which are often implemented as range queries.  Range query algorithms could be used to restrict the number of trajectories that need to be considered to a volume relevant to the expected flight path.  
\bibliography{Pre_Departure_refs}

\end{document}